\documentclass{edm_article}

\usepackage[numbers]{natbib}
\usepackage{pifont}
\usepackage{tabularx}
\usepackage{nth}
\usepackage{booktabs}
\usepackage{array}
\usepackage{amsmath}
\usepackage{multirow}
\usepackage[labelfont=bf]{caption}
\usepackage{float}
\usepackage{subcaption}
\usepackage{url}
\usepackage[dvipsnames,table]{xcolor}
\usepackage{tcolorbox}
\tcbuselibrary{skins,theorems,breakable}
\usepackage{soul}
\usepackage{arydshln}

\DeclareRobustCommand{\hlgray}[1]{{\sethlcolor{lightgray!50}\hl{#1}}}
\DeclareRobustCommand{\hlgreen}[1]{{\sethlcolor{green!25}\hl{#1}}}

\tcbset{
mybox/.style={
  breakable,
  enhanced,
  boxrule=0.4pt,titlerule=-0.2pt,drop fuzzy shadow,
  width=\linewidth,
  fonttitle=\Large\bfseries\sffamily,
  fontupper=\sffamily,
  fontlower=\sffamily,
  before=\par\medskip\noindent,
  after=\par\medskip,
  center title,
  toptitle=3pt,
  top=11pt,topsep at break=-5pt,
  frame style={left color=red!75!black,
right color=blue!75!black},
  colback=white
}}

\begin{document}
\hyphenation{CMCQRD} 

\title{Are You Doubtful? Oh, It Might Be Difficult Then!\\~Exploring the Use of Model Uncertainty for Question Difficulty Estimation}

\numberofauthors{3}
\author{
\alignauthor
Leonidas Zotos\\
       \affaddr{Center for Language and Cognition}\\
       \affaddr{Groningen, The Netherlands}\\
       \email{l.zotos@rug.nl}
\alignauthor
Hedderik van Rijn\\
       \affaddr{Department of Experimental Psychology}\\
       \affaddr{Groningen, The Netherlands}\\
       \email{d.h.van.rijn@rug.nl}
\alignauthor Malvina Nissim\\
       \affaddr{Center for Language and Cognition}\\
       \affaddr{Groningen, The Netherlands}\\
       \email{m.nissim@rug.nl}
}

\maketitle

\begin{abstract}
In an educational setting, an estimate of the difficulty of multiple-choice questions (MCQs), a commonly used strategy to assess learning progress, constitutes very useful information for both teachers and students. Since human assessment is costly from multiple points of view, automatic approaches to MCQ item difficulty estimation are  investigated, yielding however mixed success until now. Our approach to this problem takes a different angle from previous work: asking various Large Language Models to tackle the questions included in three different MCQ datasets, we leverage \textit{model uncertainty} to estimate item difficulty. By using both model uncertainty features as well as textual features in a Random Forest regressor, we show that uncertainty features contribute substantially to difficulty prediction, where difficulty is inversely proportional to the number of students who can correctly answer a question. In addition to showing the value of our approach, we also observe that our model achieves state-of-the-art results on the USMLE and CMCQRD publicly available datasets.
\end{abstract}


\section{Introduction}
\label{sec:introduction}

Multiple-Choice Questions (MCQs) are commonly used as a form of assessment across educational levels. This is not surprising, as they are trivial to grade and can effectively assess a student's knowledge, as long as they are designed well \citep{gierl_distractors2017}. Naturally, an aspect that significantly affects an MCQ's quality is its \textit{difficulty} (broadly determined by the students' success answering the question item). Intuitively, items that are too easy do not sufficiently challenge students, while very difficult items lead to frustration and demotivation impairing the learning process \citep{papouvsek2016impact}. However, estimating an item's difficulty is not trivial. In fact, students, and especially teachers, are not great at estimating how many of the test-takers will select the correct answer, given a question \citep{vandewatering2006_difficulty}. While field-testing question items is a viable solution, it is usually expensive, both in terms of time and resources. 

 \begin{figure}
    \Description{Approach overview: Predicting difficulty of Multiple-Choice Question items using textual features and uncertainty of LLM test-takers.}
    \centering
    \includegraphics[width=\linewidth]{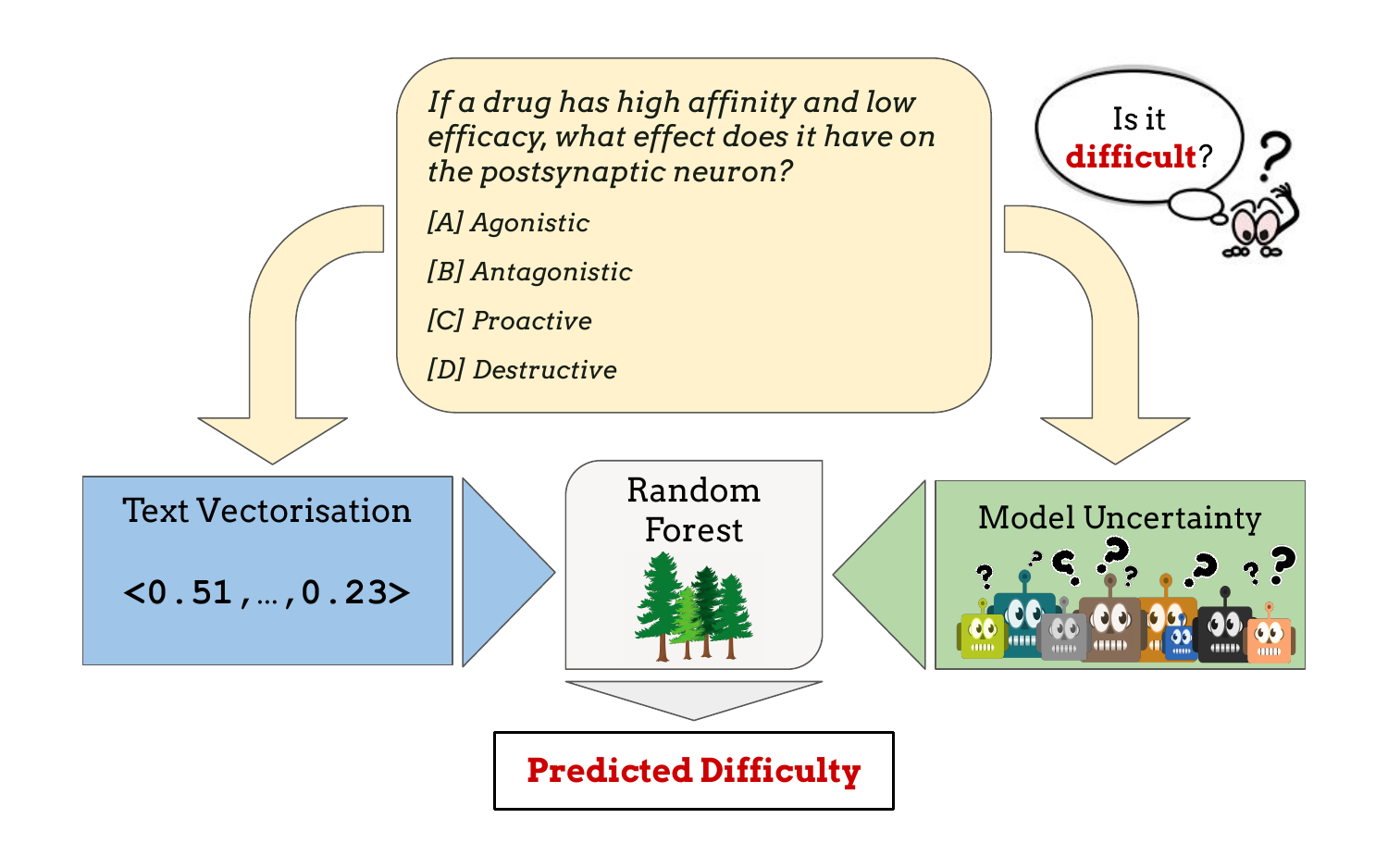}
    \caption{\textbf{Approach overview: Predicting difficulty of Multiple-Choice Question items using textual features and uncertainty of LLM test-takers.}}
    \label{fig:approach_overview}
\end{figure}

Computational methods, including Large Language Models (LLMs), have had some success in assessing the difficulty of MCQs \citep{AlKhuzaey2024}. At the same time, the task remains challenging, as shown by a recent shared task on automated difficulty prediction for MCQs \citep{yanevaFindingsFirstShared2024}, where most submitted systems performed barely above some simple baselines. The goal of the current work is to tackle the task of item difficulty estimation using a minimal experimental setup showcasing the usefulness of \textit{model uncertainty} for this task. We do this by obtaining a score for the uncertainty LLMs exhibit when answering a variety of MCQs and use it, in combination with basic text and semantic features, to train a regressor model. This expands on previous findings which showed a correlation between model and student perceived difficulty \citep{zotos-etal-2025-model}, paired with the intuition that both syntactic and semantic features are integral to this task \citep{AlKhuzaey2024}. We focus on factual MCQs, as they provide more objective assessment than open-ended questions, while still offering more complexity than True/False questions where the a baseline random chance is 50\%. In contrast, MCQs can follow various formulations and the incorrect choices play a significant role. Lastly, this choice is also motivated by dataset availability, as explained in Section~\ref{sec:methods:data}.

It is worthwhile explicitly mentioning that in the current work, the term ``uncertainty" is used to encompass both 1st token probability and choice-order probability metrics (see Section~\ref{sec:approach:model_uncertainty} for details.) These measures are taken to broadly represent the inverse of model confidence. While accurately determining the uncertainty of an LLM is an open field of research, previous research suggests that both 1st token probability \citep{plaut2024softmax} and choice order probability \citep{zotos-etal-2025-model} correlate well with model correctness in the MCQ setup. These findings also hold in the current experimental setup, as shown in Appendix~\ref{app:uncertainty_and_correctness}.

\paragraph{Our Contribution}
The contribution of our work is twofold. First, thanks to extensive experiments with a variety of LLMs and feature analysis using a regressor model (Random Forest Regressor), we showcase that model uncertainty is a useful proxy for item difficulty estimation on three different question sets assessing both factual knowledge and reading comprehension. Second, as a byproduct of our experiments investigating model uncertainty we yield a model which achieves best results to date on the BEA~2024 Shared Task dataset as well as the CMCQRD dataset. This model, together with all experimental code, is made available to the community for replicability and future extensions. We believe that our conceptual insight (model uncertainty as a useful signal for item difficulty), as well as our practical contribution in terms of an existing modular system, will foster further improvements in the task of MCQ automatic difficulty estimation, which is core in the educational setting.\footnote{Code available at: \url{https://github.com/LeonidasZotos/Are-You-Doubtful-Oh-It-Might-Be-Difficult-Then}}

\section{Related work}
The task of estimating the difficulty of MCQ items has been explored from various viewpoints in the literature \citep{AlKhuzaey2024}. Most commonly this task is tackled by training a model on a set of syntactic \citep{Perkins1995, ha-etal-2019-predicting} and/or semantic features \citep{xue-etal-2020-predicting, Hsu2018}. Furthermore, the majority of studies focus on the field of Language learning \citep{bi-etal-2021-simple-complex, he2021automatically} which is inherently different to factual knowledge or reading comprehension examinations. While the task of difficulty estimation has been widely explored, it remains challenging as was also seen in the recent ``Building Educational Applications" (BEA) shared task on ``Automated Prediction of Item Difficulty and Item Response Time", where simple baselines were overall only marginally beaten \citep{yanevaFindingsFirstShared2024}. In this task, a variety of approaches were explored with the focus ranging from architectural changes to data augmentation techniques. Notably, the best performing team (EduTec) used a combination of model optimisation techniques, namely scalar mixing, rational activation and multi-task learning (leveraging the provided response time measurements also provided in the USMLE dataset) \citep{gombert-etal-2024-predicting}. 

Most similar to our work is the study by Loginova et al. \citep{loginova2021}, who also explore the use of confidence of language models to estimate question difficulty. While similarities exist, the current research deviates considerably from this study. Whereas Loginova et al. focus on training and calibrating Encoder-Only Question-Answering models, we instead examine ``off-the-shelf" Decoder-Only models, which inherently incorporate a greater amount of factual knowledge as a byproduct of their language modeling objective \citep{zhao2023survey}. Additionally, we broaden the research scope beyond language comprehension to include factual knowledge understanding, through the use of three datasets that assess knowledge across different education levels. Finally, rather than relying on proxied comparative assessments -- where assessors classify questions as ``easy" or ``difficult" -- we leverage fine-grained, continuous difficulty labels, such as the proportion of students answering a question correctly.

More broadly, there is an emerging ``LLM-as-a-judge" field of research, which, in general terms, explores the possibility of using powerful LLMs as a substitute for human annotation \citep{zheng2023, pan2024}. For the task of question difficulty estimation, this paradigm has been explored in the context of language comprehension by with some success \citep{rainaQuestionDifficultyRanking2024}. More recently Raina et al. achieved good results in the USMLE and CMCQRD datasets through comparative assessments (i.e., given two question items, the LLM's task is to determine which is more difficult) \citep{raina-etal-2025-finetuning}. 

The present work is also influenced by the work by Zotos et al., where a variety of analyses showed a weak, but statistically significant, correlation between human and machine perceived difficulty \citep{zotos-etal-2025-model}. We take this one step further, by testing a battery of different LLMs on item difficulty estimation using their uncertainty as a signal, focusing on three distinct question sets assessing both factual knowledge and reading comprehension.

\begin{table*}[t!]
\centering
\renewcommand{\arraystretch}{1.3}  
\caption{\textbf{Examples questions from the Biopsychology, USMLE and CMCQRD datasets. Correct answer in \textcolor{ForestGreen}{green}. Two examples are given from the Biopsychology dataset to illustrate the phrasing variability.}}
\begin{tabular}{m{0.12\textwidth} m{0.55\textwidth}  m{0.23\textwidth}}
\hline
Dataset & Question & Choices \\
\hline
Biopsychology & Homeostasis is to ... as allostasis is to ... & \textcircled{a} \textcolor{ForestGreen}{constant; variable}\newline \textcircled{b} constant; decreasing\newline \textcircled{c} variable; constant \\

\noalign{\vskip 1mm}

Biopsychology & If a drug has high affinity and low efficacy, what effect does it have on the postsynaptic neuron?& \textcircled{a} agonistic \newline \textcircled{b} \textcolor{ForestGreen}{antagonistic} \newline \textcircled{c} proactive \newline \textcircled{d} destructive \\

\noalign{\vskip 1mm}  

USMLE & A 65-year-old woman comes to the physician for a follow-up examination after blood pressure measurements were 175/105 mm Hg and 185/110 mm Hg 1 and 3 weeks ago, respectively. She has well-controlled type 2 diabetes mellitus. Her blood pressure now is 175/110 mm Hg. Physical examination shows no other abnormalities. Antihypertensive therapy is started, but her blood pressure remains elevated at her next visit 3 weeks later. Laboratory studies show increased plasma renin activity; the erythrocyte sedimentation rate and serum electrolytes are within the reference ranges. Angiography shows a high-grade stenosis of the proximal right renal artery; the left renal artery appears normal. Which of the following is the most likely diagnosis? & \textcircled{a} \textcolor{ForestGreen}{Atherosclerosis} \newline \textcircled{b} {Congenital renal artery hypoplasia} \newline \textcircled{c} Fibromuscular dysplasia \newline \textcircled{d} Takayasu arteritis \newline \textcircled{e} Temporal arteritis\\

\noalign{\vskip 1mm}  

CMCQRD & Mum caught Jess by the arm.  ‘Come with me,’ she said.  Jess followed her through to the study and there on the table, propped against the wall, was an unframed painting, unmistakably one of Grandpa’s, yet unlike anything he had done before, and clearly nowhere near finished.  ‘Do you know anything about this?’ said Mum.  Jess shook her head.  ‘I’ve never seen it before.  I didn’t know he was working on anything.’ [...] And it was hard to imagine anyone, even a goddess, having any influence over someone as wilful as Grandpa.  ‘He doesn’t need me to inspire him,’ she said.  ‘He’s been painting all his life.’ \newline What impression is given of Jess’s grandfather in the final paragraph? & \textcircled{a} He lacks confidence in his ability. \newline \textcircled{b} He values Jess’s opinions. \newline \textcolor{ForestGreen}{\textcircled{c} He has a strong character.} \newline \textcircled{d} He finds it hard to concentrate. \\

\hline\end{tabular}
\label{tab:question_examples}
\end{table*}

\section{Data} 
\label{sec:methods:data}

The three MCQ datasets that we use in our experiments are described more in detail in the following subsections. The first is a dataset on the domain of Biopsychology that is not publicly available. The second is the publicly available dataset used in the BEA~2024 Shared Task~\citep{yanevaFindingsFirstShared2024}. Lastly, we also use the ``Cambridge Multiple-Choice Questions Reading Dataset" \citep{cmcqrd_2023}. For brevity, we  refer to the ``Biopsychology", ``USMLE" and ``CMCQRD" datasets respectively. Our choice is driven by the requirement of having question-sets along with students selection rates (serving as proxies for item difficulty scores). Considering that, to the best of our knowledge, the USMLE and CMCQRD datasets are the only publicly available resources satisfying this requirement. Additionally, we use a non-publicly available dataset as a complement. This choice is in line with the observation by AlKhuzaey et al., who note that most studies tackling this task resort to using private datasets \citep{AlKhuzaey2024}.

\begin{table}
    \centering
    \renewcommand{\arraystretch}{1.3}  
    \caption{\textbf{Train and Test splits as used in our experiments. For USMLE, we use the splits as provided in the competition \citep{yanevaFindingsFirstShared2024}. For the Biopsychology and CMCQRD datasets, we randomly sampled the questions, keeping the same percentage of training/testing samples as in USMLE. For the CMCQRD, we also ensured that text passages present in the train set were not present in the test set.}}
    \begin{tabular}{lcccc} 
        \hline
        Dataset & Train & Test & Total & Public?\\
        \hline
        Biopsychology & 573 & 246 & 819 & \ding{56} \\
        USMLE & 466 & 201 & 667 & \ding{52} \\
        CMCQRD & 552 & 241 & 793 & \ding{52} \\
        \hline
    \end{tabular}
    \label{tab:train_test_split}
\end{table}

As will be explained in Sections~\ref{sec:methods:data:biopsychology}~through~\ref{sec:methods:data:CMCQRD}, the three datasets vary in multiple aspects, for example question formulation, number of incorrect choices (also known as distractors) and knowledge specificity. Furthermore, to facilitate comparison with the findings from the BEA 2024 Shared Task, we use the train/test split as provided in the shared task itself (70\% training and 30\% test samples). The same proportions are also used for the Biopsychology and CMCQRD datasets, as shown in Table~\ref{tab:train_test_split}.

The item difficulty labels differ between the three datasets. In the Biopsychology and CMCQRD datasets, difficulty is measured by the proportion of students who answered correctly (a higher value indicates an easier question). This measure is commonly known as the \textit{p-value} of a question item \citep{vandewatering2006_difficulty}. In contrast, the USMLE dataset originally uses the inverse difficulty measurement, where a higher difficulty label signifies that fewer students answered correctly. Additionally, a linear transformation is also applied on the target labels of the USMLE dataset. While this difference does not affect our approach, as in both cases difficulty is conceptually expressed by cumulative student performance, to allow easier interpretation of our results we have transformed the USMLE difficulty scores to their complements such that they also reflect the proportion of correct responses per question.

\begin{figure}[h!]
    \Description{Biopsychology dataset is least balanced, with most questions being very easy. USMLE has fairly spread out p-values.}
    \centering
    \includegraphics[width=0.95\linewidth]{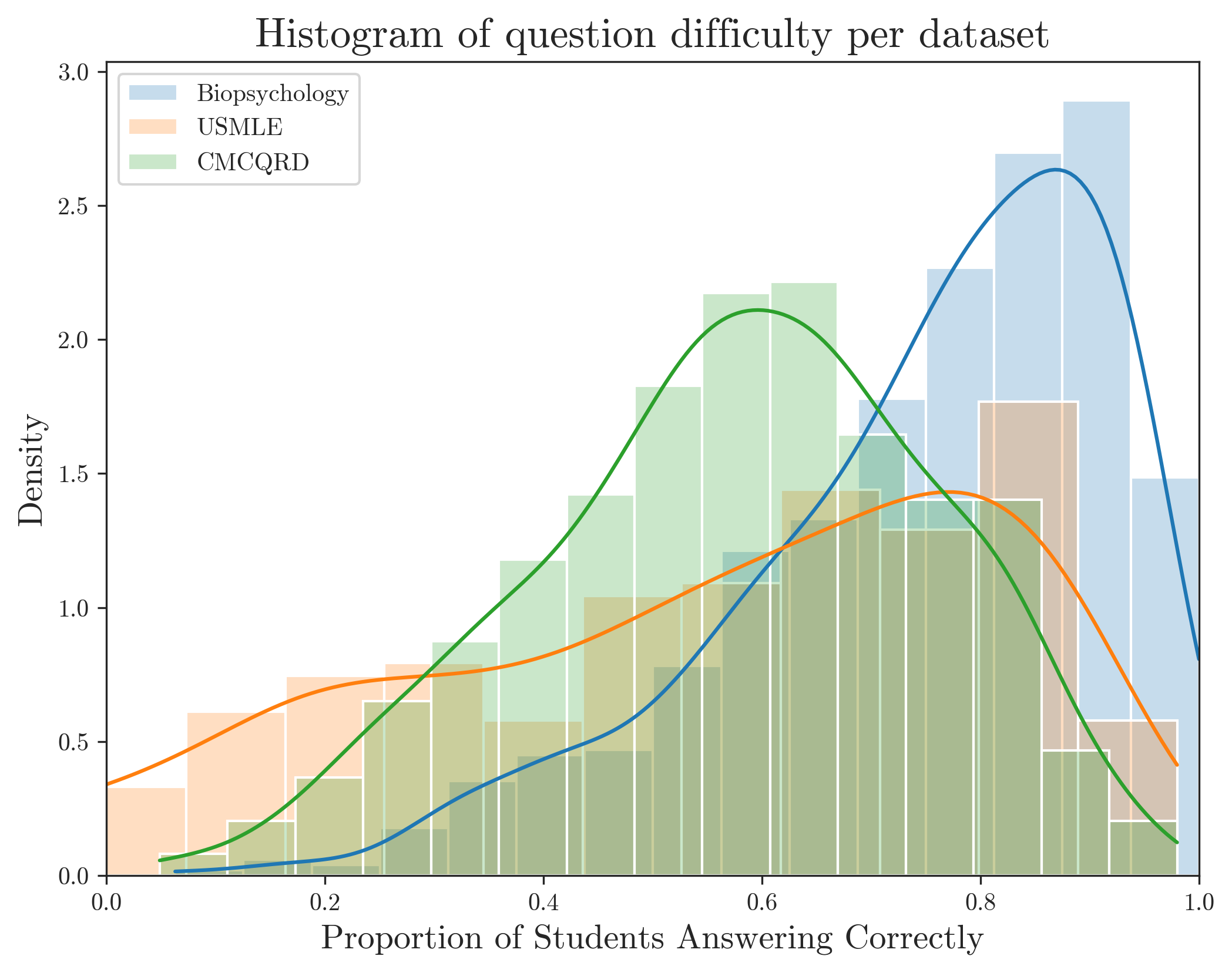}
    \caption{\textbf{Distribution of question difficulty, based on the proportion of students answering each question correctly. For the USMLE dataset, only transformed p-values are available.}}
    \label{fig:data:distribution}
\end{figure}

Finally, CMCQRD is also equipped with an Item Response Theory (IRT)-based metric, specifically scaled single-parameter Rasch Model difficulty estimates. To the best of our knowledge, while previous results exist using this IRT-based metric \citep{raina-etal-2025-finetuning}, no prior work has attempted to estimate p-values for this dataset. In this work, we conduct experiments independently estimating both the p-value and the IRT-based metric, achieving state-of-the-art results with the latter, where prior results are also available.

Figure \ref{fig:data:distribution} illustrates the distribution of difficulty across the three datasets (focusing on the p-value for the CMCQRD and not the IRT metric). As shown, while the datasets contain questions of varying difficulty levels, they are generally skewed toward easier questions. This trend is particularly evident in the Biopsychology dataset, where 81\% of questions were answered correctly by at least 60\% of the student population.

\subsection{Biopsychology}
\label{sec:methods:data:biopsychology}
The Biopsychology dataset originates from a course taught in the \nth{1} year of the Psychology undergraduate degree at a Social Sciences Faculty, covering content from the classic textbook ``Biological Psychology" by Kalat \citep{Kalat_2016}. The dataset comprises of $819$ MCQs in total, of which $451$ and $368$ have two and three distractors respectively. The data was collected from fifteen examinations with an average of $261$ examinees (Standard Deviation of $184$). This dataset has not been previously made public, minimising the risk of data contamination (ensuring that the LLMs used have not encountered the question set during training). An important feature of this question set is its high textual variability, with questions ranging from ``Fill two gaps" to ``Wh-questions". Two example questions are reported in Table~\ref{tab:question_examples}. Given that LLMs demonstrate sensitivity to input formulation \citep{bidermanLessonsTrenchesReproducible2024}, the presence of such variability in the data improves generalisation of our method across datasets.

\subsection{USMLE}
\label{sec:methods:data:USMLE}
The United States Medical Licensing Examination (USMLE) question set was developed by the National Board of Medical Examiners (NBME) and Federation of State Medical Boards (FSMB) \citep{yanevaFindingsFirstShared2024}. It consists of $667$ MCQs, each answered by more than $300$ medical school students. In contrast to the Biopsychology dataset, the questions follow strict guidelines (e.g., fixed structure, absence of misleading or redundant information in the question) and are presented with up to nine distractor choices, with the majority of the questions having five ($525$ items) or six distractors ($71$ items). An example instance is provided in Table~\ref{tab:question_examples}. As can be seen, questions of this dataset are longer ($755$ characters compared to $103$ characters for the Biopsychology set) and are of technical nature.

\subsection{CMCQRD}
\label{sec:methods:data:CMCQRD}
The Cambridge Multiple-Choice Questions Reading Dataset (CMCQRD) consists of 120 text passages, each accompanied with a number of MCQs (total of 793 question items) aiming at evaluating the student's language comprehension abilities. While the questions target various Common European Framework of Reference for Languages (CEFR) proficiency levels (B1 to C2), we do not leverage this additional information in our experiments. Furthermore, compared to the Biopsychology and USMLE datasets, question items in the CMCQRD are the longest, with an average of $3,618$ characters per item.

\section{Approach}
Given an MCQ, the task is to predict its difficulty measured by the proportion of students that select the correct choice (in addition to the IRT-based metric available for the CMCQRD dataset). An MCQ item consists of the stem/question, a single correct choice/answer and a number of incorrect choices/distractors (also known as ``foils").

Figure \ref{fig:approach_overview} illustrates our approach to this task. Our design is centered around a simple Random Forest Regressor\footnote{As provided by the Scikit-Learn Library, using the default hyper-parameters \citep{scikit-learn}.} which receives as input a vectorised representation of the MCQ, as well the uncertainty of multiple LLMs answering the same MCQ\footnote{Simple vector concatenation is used to combine the text and uncertainty features.}. We opted for a relatively simple Random Forest Regressor, as it allows for analysis using explainability methods (through SHAP post-hoc analysis), while still effectively demonstrating the usefulness of model uncertainty in this context. As features, we use Textual Features and Model Uncertainty, as described in sections \ref{sec:approach:textual_features} and \ref{sec:approach:model_uncertainty}. We further supplement this basic system using higher-level textual features namely Linguistic Features and Choice Similarity, described in section \ref{sec:approach:additional_features}.

\subsection{Textual Features}
\label{sec:approach:textual_features}
Intuitively, extracting the semantic content of the question item is integral to assess its difficulty. To accomplish that, we use two fundamentally different methods -- Term Frequency-Inverse Document Frequency (TF-IDF) Scores and Semantic Embeddings -- to encode the question and answer choices as numerical vectors.

\paragraph{TF-IDF Scores}
TF-IDF Scores capture how important a word is to a document within a collection by balancing its frequency in that specific document against its rarity across all documents \citep{sparck1972statistical}. In the current context, we consider each question item (along with its choices) as a single document. To capture multi-word technical terms, such as ``interstitial fibrosis", our analysis considers both individual words (unigrams) and two-word combinations (bigrams). Furthermore, we disregard terms that appear in more than 75\% of documents, and only use the $1000$ most important features (as determined by the TF-IDF values) to increase efficiency.

\paragraph{Semantic Embeddings}
Word embeddings are a technique whereby words are encoded as dense vectors in a continuous vector space, capturing semantic relationships between words. We evaluate two embedding approaches: General BERT embeddings \citep{devlin-etal-2019-bert} and domain-specific Bio-Clinical BERT embeddings \citep{alsentzer-etal-2019-publicly}. The Bio-Clinical BERT embeddings, previously also employed by team ITEC in the BEA 2024 shared task \citep{tack-etal-2024-itec}, offer specialized medical domain text encoding that potentially encapsulates more accurately the semantic content of each question item. Both techniques yield a 768-dimensional vector representation. We only conduct experiments using Bio-Clinical BERT embeddings in the Biopsychology and USMLE datasets, as CMCQRD does not focus on the biology/medical domain.

\subsection{Model Uncertainty}
\label{sec:approach:model_uncertainty}
The current methodological approach is founded on the premise that model uncertainty correlates with student performance and thus, by extension, offers a useful signal when estimating the difficulty of a question item. To explore this hypothesis, we have conducted experiments using two metrics that are shown to correlate well with model correctness (as discussed in Appendix ~\ref{app:uncertainty_and_correctness}): \textit{1st Token Probability} and \textit{Choice Order Sensitivity}. These uncertainty scores are obtained  for each LLM separately and concatenated into a single vector, to which any additionally textual, linguistic or choice similarity features are (optionally) also added. This vector is then fed to the regressor.

\paragraph{1st Token Probability}
The first technique to measure model uncertainty is by inspecting the softmax probability of the 1st token to be generated as the answer id to the given MCQ question, (e.g., probability of generating token ``B"), in comparison to the probabilities of the alternatives (e.g.,  probability of generating token ``A" or ``C"). As the 1st token probabilities can be influenced by the order in which the choices are provided in the problem set \citep{wei2024unveiling, wang2023large, wang2024look, zheng2024large}, we create ten random different orderings for each question and let the model answer each MCQ ten times\footnote{For questions with only 3 choices, we instead consider all six different choice orderings.}. This way, we calculate the average probability per MCQ choice. We then consider the average probability for the correct answer as the uncertainty metric of the LLM. 

Furthermore, as different tokens might be generated to represent the same answer (e.g., ``A", `` A", ``a ", see details on prompting and answer elicitation in Section~\ref{sec:methods:choice_of_models_and_prompting} below) and different models might attribute higher likelihood for specific tokens, the token representing each choice with the highest probability is selected. For example if for a given model the probability of generating token ``C" is higher than the probability of token ``c", the former is considered for that model. Lastly, the three extracted mean probabilities of all orderings are normalised in the range of $0$ to $1$.

\paragraph{Choice Order Sensitivity}
Pezeshkpour and Hruschka \citep{pezeshkpour2023large} observed that choice order sensitivity correlates with error rate. In other words, when LLMs consistently select a choice regardless of its position, that choice is more likely to be correct. Based on this observation, we leverage this correlation to measure  uncertainty. Specifically, for all evaluated choice orderings, we measure the probability of the correct choice being selected. Thus, this probability is not based on token probabilities but rather on the eventual choice.

\subsection{Additional Features}
\label{sec:approach:additional_features}
We also conduct experiments with two additional feature sets: linguistic and choice similarity features. First, based on the work of Ha et al. \citep{ha-etal-2019-predicting}, we extract 17 higher-level linguistic features from each question item. These features vary in complexity, ranging from the number of sentences to the occurrence of additive connectives. Moreover, we also develop a simple baseline that uses these linguistic features along with \texttt{word2vec} embeddings \citep{word2vec}, similar to the work by Yaneva et al. \citep{yaneva-etal-2021-using}\footnote{It is worth noting that in contrast to Yaneva et al. \citep{yaneva-etal-2021-using}, we do not use readability metrics as some question items do not meet the $100$ word requirement for these metrics to be computed}. 

In the current work we also explore the use of choice similarity defined (per MCQ) as the average cosine similarity between each distractor and the correct answer choice, similar to the approach by \citep{susanti2020integrating}. This is operationalised using the Sentence Transformer library \citep{reimers-2019-sentence-bert} with one of two models: \texttt{all-MiniLM-L6-v2}\footnote{https://huggingface.co/sentence-transformers/all-MiniLM-L6-v2} (general embeddings) and \texttt{S-PubMedBert-MS-MARCO}\footnote{\url{https://huggingface.co/pritamdeka/S-PubMedBert-MS-MARCO}} (medical/health text domain embeddings). The two setups are henceforth referred to as ``General Similarity" and ``Medical Similarity" respectively.

\subsection{Choice of Models and Prompting}
\label{sec:methods:choice_of_models_and_prompting}
In this work, we focus on Decoder-Only models, as they are considered to have incorporated greater amounts of factual knowledge as a byproduct of their language modeling objective \citep{zhao2023survey}, compared to Encoder-Only or Encoder-Decoder models. Moreover, as the internal logit probabilities of the 1st token to be generated are needed to measure the uncertainty of each model, we focus on nine open-sourced models of different parameter sizes and families. Additionally, we use the models' instruction-tuned variant and experiment with both default precision and 4-bit quantised models. This comparison is important because, although quantized models are more efficient (albeit slightly less capable), it is unclear whether they are also more miscalibrated\footnote{Details of the models used are in Table~\ref{tab:LLMs_used} in Appendix~\ref{app:LLMs_used}.}. To adapt them for the task of MCQ answering, we use the instruction prompt in Figure~\ref{tab:instruction_phrasing} based on the experimental setup of Plaut et al. \citep{plaut2024softmax} as well as Zotos et al. \citep{zotos-etal-2025-model}. 

\begin{figure}[h!]
    \Description{Instruction phrasing used for all models and experiments: Below is a multiple-choice question. Choose the letter which best answers the question. Keep your response as brief as possible; just state the letter corresponding to your answer with no explanation.}
    \centering
        \tcbset{colback=white,colframe=black,
        fonttitle=\bfseries}
        \begin{tcolorbox}[enhanced,title=Instruction Prompt for the LLM,
            frame style={color=gray}
        ]
        Below is a multiple-choice question. Choose the letter which best answers the question. Keep your response as brief as possible; just state the letter corresponding to your answer with no explanation. \newline
        
        Question: \newline \textit{[Question Text]}
        
        Response:
        \end{tcolorbox}
    \caption{\textbf{Instruction phrasing used for all models and experiments. \textit{[Question Text]} is replaced by the item stem followed by the answer choices, each prepended with the corresponding letter \textit{A} to \textit{J}.}}
    \label{tab:instruction_phrasing}
\end{figure}

\begin{table*}[h!]
\centering
\renewcommand{\arraystretch}{1.2}  
\caption{\textbf{Root Mean Squared Error (RMSE, the lower the better) on the test set using different sets of features. Lowest achieved RMSE per dataset is shown in \textbf{boldface}. Best overall results are highlighted in \hlgreen{light green}.  All results are averaged over ten repetitions, with the standard deviation not exceeding 0.002.}}
\label{tab:results:main_results}
\begin{tabular}{lcc:cc:cc:cc} 
\hline
\textbf{Method} & \multicolumn{2}{c}{\textbf{Biopsychology}} & \multicolumn{2}{c}{\textbf{USMLE}} & \multicolumn{2}{c}{\textbf{CMCQRD}} & \multicolumn{2}{c}{\textbf{CMCQRD\_IRT}} \\  
\hline
\multicolumn{7}{l}{\hlgray{\textbf{Baselines}}} \\
Dummy Regressor                                             & \multicolumn{2}{c}{0.1667}  & \multicolumn{2}{c}{0.3110}  & \multicolumn{2}{c}{0.1833} & \multicolumn{2}{c}{9.7568}      \\
Best Literature Result \citep{raina-etal-2025-finetuning}   & \multicolumn{2}{c}{-}       & \multicolumn{2}{c}{0.291 }  & \multicolumn{2}{c}{-} & \multicolumn{2}{c}{8.5}\\
\hline
\multicolumn{7}{l}{\hlgray{\textbf{Only Text}}} \\
TF-IDF                                                      & \multicolumn{2}{c}{0.1479}  & \multicolumn{2}{c}{0.3092}  & \multicolumn{2}{c}{0.1843} & \multicolumn{2}{c}{9.2531}      \\
BERT Embeddings                                             & \multicolumn{2}{c}{0.1498}  & \multicolumn{2}{c}{0.3066}  & \multicolumn{2}{c}{0.1843} & \multicolumn{2}{c}{8.9347}      \\  
\hline
\multicolumn{7}{l}{\hlgray{\textbf{Only Uncertainty}}} \\
1st Token Probabilities                                     & \multicolumn{2}{c}{0.1473}  & \multicolumn{2}{c}{0.3041}  & \multicolumn{2}{c}{0.1770} & \multicolumn{2}{c}{9.4161}     \\
Choice Order Sensitivity                                    & \multicolumn{2}{c}{0.1543}  & \multicolumn{2}{c}{0.3155}  & \multicolumn{2}{c}{0.1788} & \multicolumn{2}{c}{9.9787}     \\
Both Uncertainty Features                                   & \multicolumn{2}{c}{0.1460}  & \multicolumn{2}{c}{0.3034}  & \multicolumn{2}{c}{0.1754} & \multicolumn{2}{c}{9.3705}     \\  
\hline
\multicolumn{7}{l}{\hlgray{\textbf{Text and Uncertainty}}} \\
                                                            & \textbf{TF-IDF}   & \textbf{BERT}  & \textbf{TF-IDF} & \textbf{BERT}   & \textbf{TF-IDF}      & \textbf{BERT}      & \textbf{TF-IDF} & \textbf{BERT}        \\
First Token Probability                                     & 0.1361            & 0.1362         & 0.3037          & 0.2868          & 0.1668               & 0.1669             & 8.5661          & \textbf{8.2763}      \\
Choice Order Sensitivity                                    & 0.1378            & 0.1409         & 0.310           & 0.2906          & 0.1658               & 0.1676             & 8.6663          & 8.4204               \\
Both Uncertainty Features                                   & \hlgreen{\textbf{0.1359}}  & 0.1365         & 0.3044          & \textbf{0.2864} & \textbf{0.1654}      & 0.1669             & 8.5933          & 8.3009               \\  
\hline
\end{tabular}
\end{table*}

\section{Results}
\label{sec:results}

Our experiments are aimed at evaluating the usefulness of model uncertainty as a signal for MCQ item difficulty as well as discovering which specific textual and uncertainty features are most relevant for our trained Regressor. We first focus on the performance of our setup using text and model uncertainty feature sets (Section~\ref{sec:results:performance_on_item_difficulty_estimation}), followed by an exploration of the effect of additional features (Section~\ref{sec:results:effect_of_additional_features}). We conclude with a post-hoc analysis of our system using SHAP explanations (Section~\ref{sec:results:feature_importance}). While the following sections present only the relevant results, Appendix~\ref{app:all_results} provides an overview of the results for all experiments. All experiments were conducted using two Nvidia A100 GPUs.

\subsection{Performance on Difficulty Estimation}
\label{sec:results:performance_on_item_difficulty_estimation}
To evaluate the performance of our trained models we use the Root Mean Squared Error (RMSE) metric from Python’s Scikit-learn library~\citep{scikit-learn}, as used in the BEA 2024 Shared task. As previously mentioned, we use a Random Forest Regressor tasked to predict the difficulty of a question item, given as input a vectorised representation of the MCQ as well as the uncertainty of multiple LLMs answering the same MCQ. This creates a modular setup that allows easy manipulation of the input feature set. We present the feature sets along with their performance on the three datasets in Table~\ref{tab:results:main_results}. For brevity, we report the results obtained using Bio-Clinical BERT Embeddings in Appendix~\ref{app:all_results}, as they were found to lead to similar, yet consistently slightly worse RMSE scores compared to the general BERT embeddings. Similarly, Table~\ref{tab:results:main_results} presents the performance of the default precision variants of the models, while the results for the quantised models are reported in Appendix~\ref{app:all_results}. Lastly, for CMCQRD, we report the performance on predicting both p-values and the IRT-based metric.

An important first observation is that the RMSE difference between experiments is minimal. This is in-line with the findings from the BEA 2024 shared task, where the lowest achieved RMSE was only $0.012$ lower than the baseline, and the achieved RMSE scores of the top 10 approaches were within 0.009. Even so, there are consistent differences between the experimental setups. Most importantly for this research, incorporating model uncertainty alongside text features significantly reduces RMSE across all datasets, outperforming the best scores in previous literature. Even lower RMSE is achieved for the USMLE and CMCQRD datasets when additional features are included (see Section~\ref{sec:results:effect_of_additional_features}). Furthermore, our exploration revealed that 1st Token Probability consistently serves as a more useful signal for the task than Choice Order Sensitivity. However, combining both yields the best results. Regarding the two text vectorization methods, we find no significant differences between them, except for the USMLE dataset, where BERT embeddings outperform TF-IDF scores.

\subsection{Effect of Additional Features}
\label{sec:results:effect_of_additional_features}
Naturally, additional features of an MCQ might capture aspects that make the question easier or more challenging. Table \ref{tab:similarity_and_ling_features} explores the effect of higher-level linguistic features and the similarity between choices (as described in Section~\ref{sec:approach:additional_features}). Expectedly, using choice similarity as the only feature yields poor results on the task. Similarly, the baseline that focuses on linguistic features only marginally beats the Mean Regressor baseline for two of the four experiments. However, combining either, or both, of these features with text and model uncertainty features further improves our best results for all but the Biopsychology experiments (where the best result is still achieved without the additional features). Moreover, we observe that there is no clear advantage between the General or Medical Similarity, with the latter also seemingly being useful in the non-medical domain (CMCQRD dataset). These experiments highlight that while model uncertainty and basic text encodings (such as BERT or TF-IDF scores) capture certain aspects of difficulty, other factors should still be utilized for this task.

\begin{table*}[h!]
\centering
\renewcommand{\arraystretch}{1.2}  
\caption{\textbf{Root Mean Squared Error (RMSE, the lower the better) on the test set using additional features. Lowest achieved RMSE per dataset is shown in \textbf{boldface}. Best overall results, also in comparison to those presented in Table~\ref{tab:results:main_results}, are additionally highlighted in \hlgreen{light green}. All results are averaged over ten repetitions, with the standard deviation not exceeding 0.002.}}
\label{tab:similarity_and_ling_features}
\begin{tabular}{lcc:cc:cc:cc} 
\hline
\textbf{Method} & \multicolumn{2}{c}{\textbf{Biopsychology}} & \multicolumn{2}{c}{\textbf{USMLE}} & \multicolumn{2}{c}{\textbf{CMCQRD}} & \multicolumn{2}{c}{\textbf{CMCQRD\_IRT}} \\  
\hline
\multicolumn{7}{l}{\hlgray{\textbf{Baselines}}} \\
Dummy Regressor                                             & \multicolumn{2}{c}{0.1667}  & \multicolumn{2}{c}{0.3110}  & \multicolumn{2}{c}{0.1833} & \multicolumn{2}{c}{9.7568}      \\
Best Literature Result \citep{raina-etal-2025-finetuning}   & \multicolumn{2}{c}{-}       & \multicolumn{2}{c}{0.291}  & \multicolumn{2}{c}{-} & \multicolumn{2}{c}{8.5}\\
Linguistic Features Baseline \citep{ha-etal-2019-predicting}& \multicolumn{2}{c}{0.1544} & \multicolumn{2}{c}{0.3147}  & \multicolumn{2}{c}{0.1852} & \multicolumn{2}{c}{9.3335}      \\
\hline
\multicolumn{7}{l}{\hlgray{\textbf{Only Choice Similarity}}} \\
General (\texttt{all-MiniLM-L6-v2})                        & \multicolumn{2}{c}{0.1895}         & \multicolumn{2}{c}{0.3567}       & \multicolumn{2}{c}{0.2226}               & \multicolumn{2}{c}{12.2829}               \\
Medical (\texttt{S-PubMedBert-MS-MARCO})                   & \multicolumn{2}{c}{0.1883}         & \multicolumn{2}{c}{0.3432}       & \multicolumn{2}{c}{0.2146}               & \multicolumn{2}{c}{11.4768}               \\  
\hline
                                                           & \textbf{TF-IDF}   & \textbf{BERT}  & \textbf{TF-IDF} & \textbf{BERT}   & \textbf{TF-IDF}      & \textbf{BERT}      & \textbf{TF-IDF} & \textbf{BERT}        \\
\multicolumn{7}{l}{\hlgray{\textbf{Text, Both Uncertainties \& Choice Similarity}}} \\
General Similarity                                         & 0.1372            &  0.1367        &  0.2835         & 0.2862          & \hlgreen{\textbf{0.1651}}   & 0.1669             & 8.5934          &  8.2956                 \\
Medical Similarity                                         & 0.1376            &  \textbf{0.1361} &  0.2836         & 0.2853        & \hlgreen{\textbf{0.1651}}   & 0.1664             & 8.5999          &  \hlgreen{\textbf{8.2325}}        \\
Both Similarities                                          & 0.1378            &  0.1365        &  0.2847         & 0.2862          & 0.1653            & 0.1676             & 8.5694          &  8.2817                 \\  
\hline
\multicolumn{7}{l}{\hlgray{\textbf{Text, Both Uncertainties, Both Sim \& Linguistic Features}}}  \\
       & 0.1393            &  0.1362        &  \hlgreen{\textbf{0.2817}}            & 0.2857          & 0.1652               & 0.1673             & 8.5490          &  8.6575             \\
\hline
\end{tabular}
\end{table*}

\subsection{Feature Importance}
\label{sec:results:feature_importance}
In order to better understand which features drive the predictions of the Random Forest Regressor, we use Shapley additive explanations as provided by the SHAP Python library \citep{shap2017}. To maintain conciseness, we present SHAP summary plots for a selected subset of experiments that we found to be the most insightful.

\begin{figure}[h!]
    \Description{SHAP summary plot showing the contribution of the top ten uni/bi-gram features to the Random Forest's predictions}
    \centering
    \includegraphics[width=\linewidth]{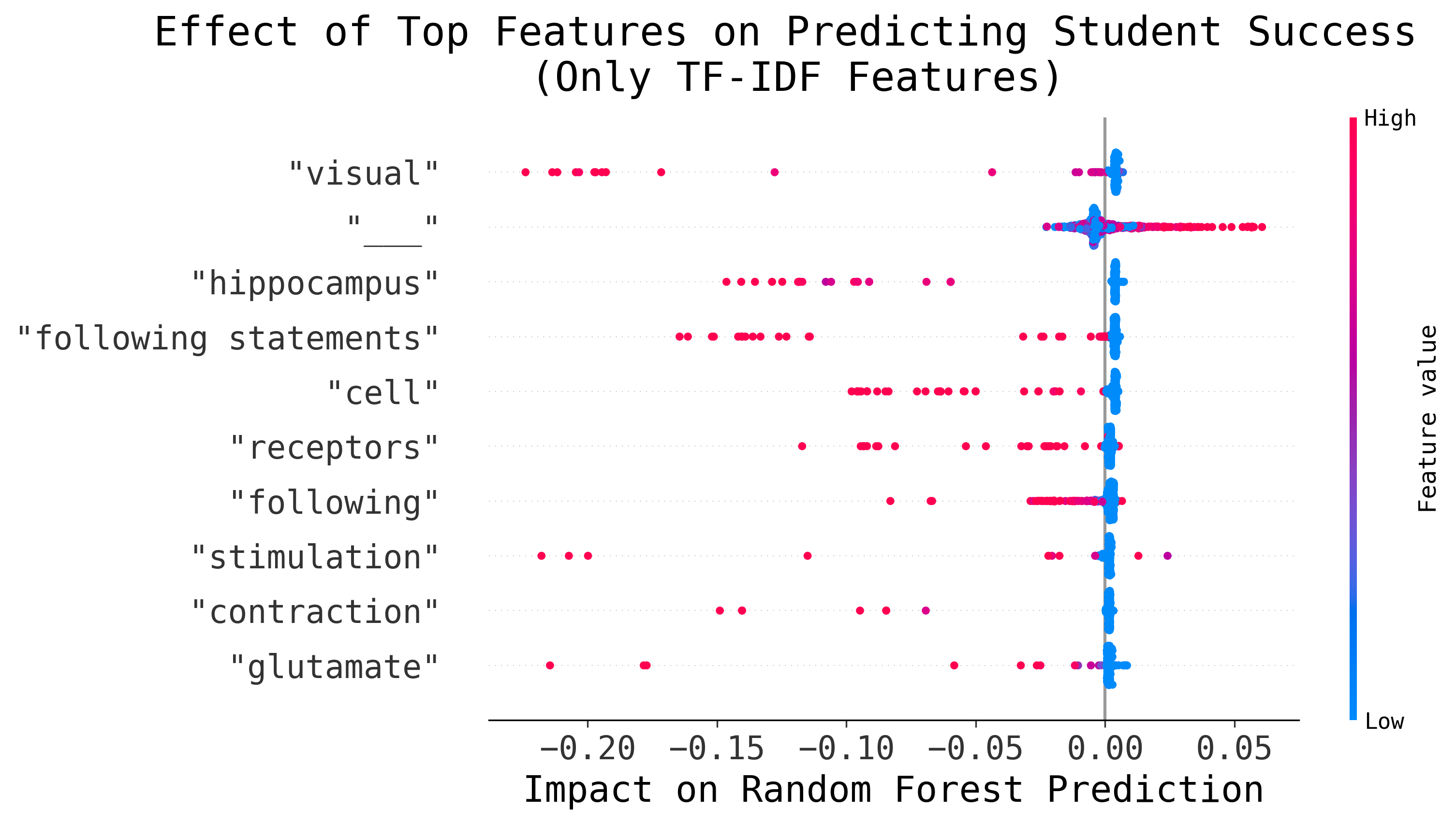}
    \caption{\textbf{Biopsychology Dataset. Shapley summary plot showing the contribution of the top ten uni/bi-gram features to the Random Forest's predictions, highlighting their importance and impact direction. Features are ranked by their average influence, with dots representing individual question items and colour indicating TF-IDF scores. Results averaged over ten repetitions.}}
    \label{fig:results:bio_shap_only_tf_idf}
\end{figure}

\begin{figure}[h!]
    \Description{Model Uncertainty has a high effect on prediction. Some words are also influential, but the most influence is from a variety of models}
    \centering
    \begin{subfigure}{\linewidth}
        \centering
        \includegraphics[width=\linewidth]{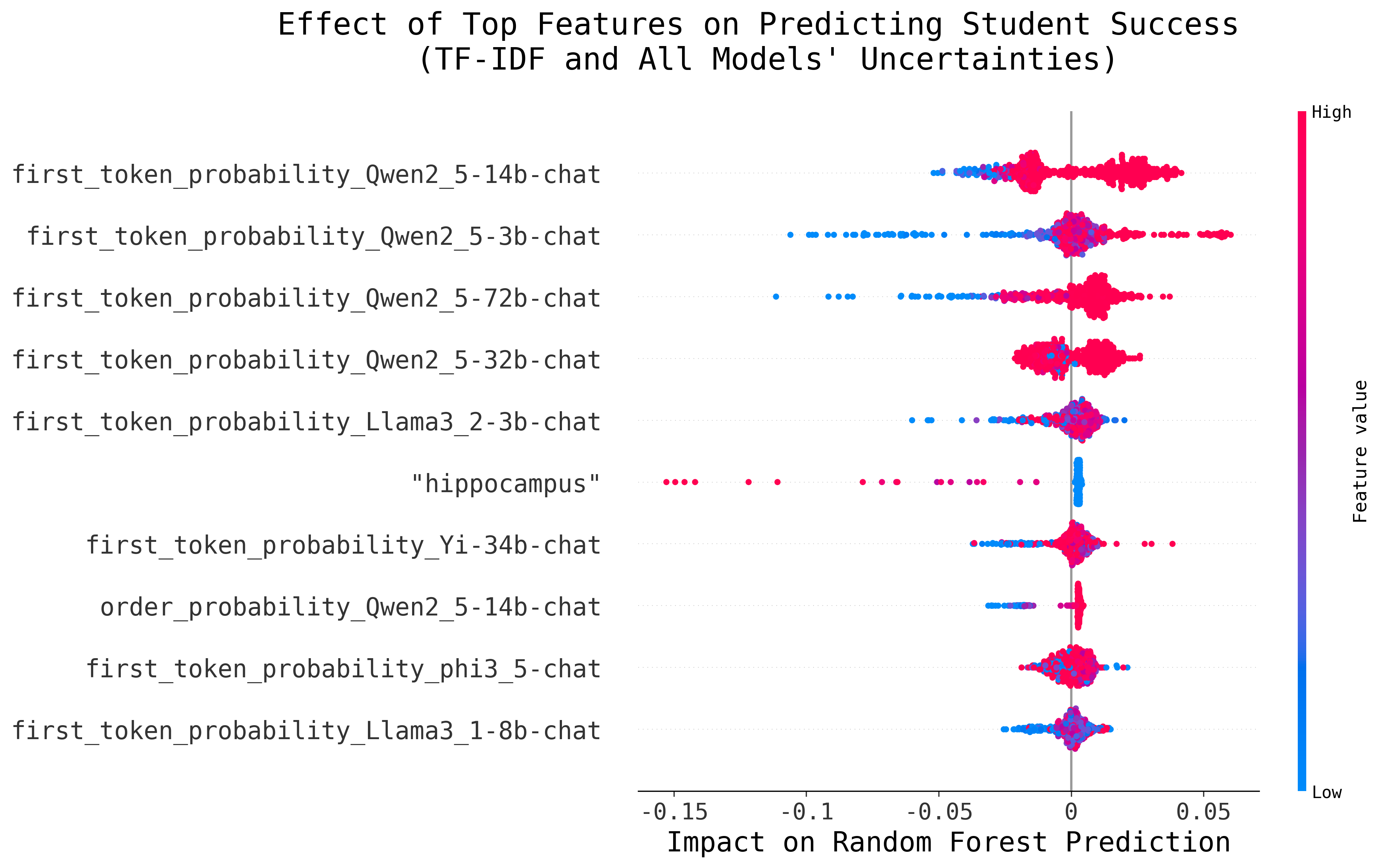}
        \caption{\textbf{Biopsychology}}
        \label{fig:results:bio_shap_best_setup}
    \end{subfigure}
    \vskip 2mm
    \begin{subfigure}{\linewidth}
        \centering
        \includegraphics[width=\linewidth]{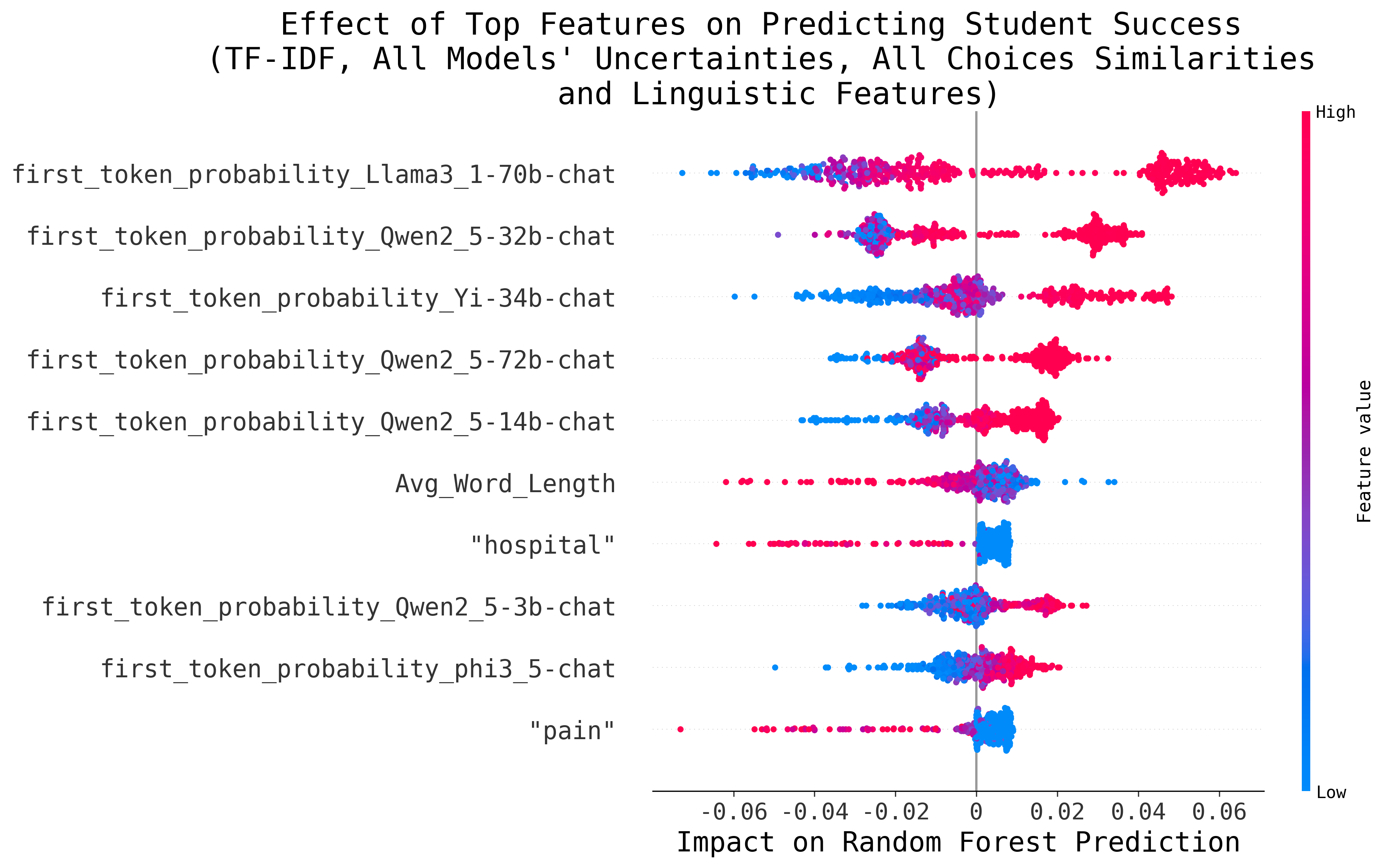}
        \caption{\textbf{USMLE}}
        \label{fig:results:usmle_shap_best_setup}
    \end{subfigure}
    \vskip 2mm
    \begin{subfigure}{\linewidth}
        \centering
        \includegraphics[width=\linewidth]{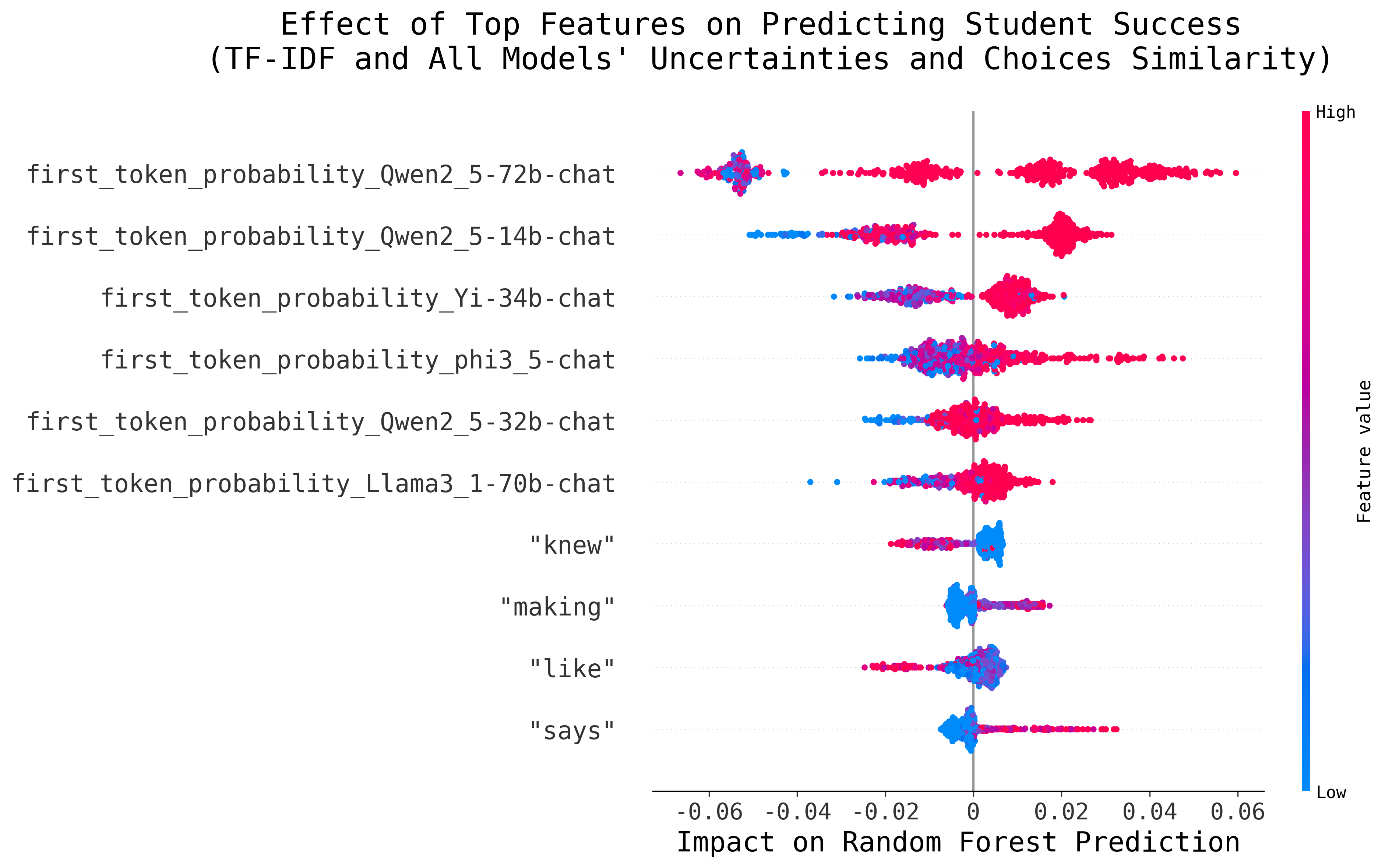}
        \caption{\textbf{CMCQRD (p-value)}}
        \label{fig:results:cmcqrd_shap_best_setup}
    \end{subfigure}
    \caption{\textbf{Shapley summary plots for the three datasets showing the contribution of the top ten features to the Random Forest's predictions. Higher First Token Probability and Order Probability metrics indicate greater model certainty. Results averaged over ten repetitions.}}
    \label{fig:results:shap_comparison}
\end{figure}

Before exploring the analysis regarding model uncertainty, we examine the contribution of the most impactful uni/bi-grams from the text-only experiment using the Biopsychology dataset. This is useful because it allows us to gain an overview of the influence of lexical features before introducing model uncertainty, while also highlighting any features that unexpectedly influence question difficulty (e.g., interrogative words).

This analysis relies on TF-IDF scores, as BERT embeddings cannot directly be traced back to individual words. Figure~\ref{fig:results:bio_shap_only_tf_idf} shows the ten most impactful features, along with their effect on the Regressors' prediction for each MCQ item. High TF-iDF scores are highlighted in red and broadly represent the presence of the word in an item. Furthermore, the impact on the Regressor's prediction is either positive or negative, meaning that a feature can either lower or increase the predicted difficulty score. As can be seen, the presence of certain terms (e.g., ``visual", ``hippocampus") lead the Regressor to predict higher difficulty. Interestingly, this analysis also demonstrates that questions where a gap (represented by an underscore ``\_") needs to be filled (e.g., ``fill-the-gap" or sentence completion) are predicted to be easier.

While this analysis shows that the presence of certain words can steer the Regressor towards predicting a higher or lower difficulty, we are mostly interested in the contribution of features related to model uncertainty. Figures~\ref{fig:results:bio_shap_best_setup}, ~\ref{fig:results:usmle_shap_best_setup}~and~\ref{fig:results:cmcqrd_shap_best_setup} present the effect of the most impactful features for the feature sets that lead to the best performance (using TF-IDF scores for the text encoding and model uncertainty) for the Biopsychology, USMLE and CMCQRD datasets, respectively. It is worth noting that for the CMCQRD dataset, we focus on the prediction of the number of students selecting the correct rate (p-value), instead of the IRT-based metric.

In all instances, the Random Forest Regressor heavily relies on model uncertainties to predict item difficulty. As hypothesised, the higher the model certainty (in terms of either 1st Token or Choice-Order Probability) the more students are predicted to answer the question correctly. In each configuration, the uncertainty of different models has the greatest influence. Notably, the uncertainty of Qwen models consistently serves as a strong indicator of difficulty. Furthermore, this analysis hints towards model size being important, especially when comparing the Biopsychology and the USMLE results: For the latter, the confidence of larger models is more influential in the Regressor's prediction of question difficulty. This observation also highlights the core challenge of our approach: having a model that is sufficiently capable of answering the MCQs but not so complex that it answers them with complete confidence. In our work, this challenge is partially addressed using an ensemble of models, leaving it up to the Random Forest Regressor to determine their usefulness.

\section{Discussion and Conclusion}
We explored how model uncertainty can be leveraged for the task of question item difficulty estimation using three MCQ datasets focusing on factual knowledge and language comprehension. We demonstrate, in experimental setups of varying complexity, that while both textual features (e.g., encoding using TF-IDF Scores or BERT embeddings) and model uncertainty features are useful for the task, the trained Random Forest Regressor performed significantly better when model uncertainty features were included.

Our results suggest that aspects of a question item that challenge students similarly impact LLMs. A factor that could explain this alignment is representation: Knowledge that is well represented in an LLM's training data is likely to be more foundational (e.g., ``What is a neuron"), compared to specialised knowledge (e.g., a medical diagnosis). By extension, using model uncertainty for this task requires a model of appropriate size/capabilities. Additionally, our results suggest that an LLM's uncertainty does not fully account for certain aspects of item difficulty, such as linguistic complexity\footnote{Notably, a question item can be complex yet easy, or vice versa \citep{pelanek_difficulty_complexity}.}. 

Our methodological design is intentionally simple, serving as a proof of concept for this approach. This simplicity stems from various design choices. Firstly, we use a variety of LLMs without placing great emphasis on their uncertainty behaviour. Specifically, while we ensure that the measured model uncertainty aligns with model correctness (as shown in Appendix~\ref{app:uncertainty_and_correctness}), we do not focus on calibrating the LLMs. Instead, we rely on the Random Forest Regressor to select and weight the uncertainties of the various models.  Secondly, we conducted a series of experiments using features of varying complexity to demonstrate that, while incorporating additional features (e.g., average cosine similarity between each distractor and the correct answer choice) can improve task performance, the improvements are often marginal compared to relying solely on a simple text encoding combined with model uncertainty.

Shifting into a broader perspective, our findings suggest that there are similarities between the way LLMs and students process educational material. While caution is necessary, we see potential for future research to leverage LLMs for student and cohort modeling.

\section*{Limitations}
\label{sec:limitations}
Indisputably, the central limitation of our approach is the reliance on (un)certain LLMs. As seen in Section \ref{sec:results}, model uncertainty is beneficial only when the model can answer the question without being overly confident. Naturally, this limits the usefulness of our approach, especially given the rapid development of LLMs in terms of their capabilities. We hypothesise that this limitation can at least partially be resolved by using calibrated LLMs, which we leave for future work.

Similarly, our approach is not expected to perform as well on MCQs designed to test knowledge at lower education levels (e.g., primary school geography exams), as even small LLMs are now capable of confidently answering such questions. At the same time, using less proficient LLMs introduces different challenges, particularly regarding linguistic ability: Smaller LLMs are more strongly affected by linguistic perturbations (e.g., question formulation, choice order) and have greater limitations in instruction-following capabilities \citep{bidermanLessonsTrenchesReproducible2024, sclar2023quantifying}.

Lastly, due to dataset availability, we evaluated our approach solely on three examinations. It remains unclear whether model uncertainty could also help assess the difficulty of exams in other skill sets, such as mathematical reasoning.

\bibliographystyle{abbrv}
\bibliography{sigproc} 
\appendix
\section{Model Correctness and Uncertainty}
\label{app:uncertainty_and_correctness}
Table \ref{tab:uncertainty_and_correctness} presents the performance of each model on the three question sets, as well as the relation between model certainty and model correctness. In line with the results of Plaut et al.  \citep{plaut2024softmax}, it is clear that both tested metrics correlate well with model correctness: On average, the mean certainty for the chosen option is higher for the correctly answered question items. This suggests that the two metrics indeed capture an aspect of model certainty. Lastly, we can see that this trend persists despite quantisation, although quantised models generally exhibit lower performance compared to their default precision counterparts.

\begin{table*}[h!]
\centering
\renewcommand{\arraystretch}{1.2}  
\caption{\textbf{Model correctness and answer probability in terms of Mean 1st Token and Choice Order Probability in the Biopsychology, USMLE and CMCQRD question sets. ``Overall Correctness" indicates the proportion of correctly answered questions, while the probabilities in \textcolor{ForestGreen}{green} and \textcolor{BrickRed}{red} indicate the mean model certainty for the model's choice for \textcolor{ForestGreen}{correctly} and \textcolor{BrickRed}{incorrectly} answered questions respectively. As can be seen, on average, model certainty is higher when questions are answered correctly, especially for larger LLMs.}}
\label{tab:uncertainty_and_correctness}
\begin{tabular}{ccccccccc}\hline 
 &  & \multicolumn{3}{c}{Default Precision} & \multicolumn{3}{c}{4-bit Quantisation} \\
\cmidrule(lr){3-5} \cmidrule(lr){6-8}
\multirow{2}{*}{Dataset} & \multirow{2}{*}{Model} & \multirow{2}{*}{Correctness} & \multicolumn{2}{c}{Mean Probability} & \multirow{2}{*}{Correctness} & \multicolumn{2}{c}{Mean Probability}  \\
\cmidrule(lr){4-4} \cmidrule(lr){5-5} \cmidrule(lr){7-7} \cmidrule(lr){8-8} 
& & & 1st Token & Choice Order & & 1st Token & Choice Order \\ \hline
\multirow{9}{*}{Biopsychology} & phi3\_5 & 0.821 & \textcolor{ForestGreen}{0.921} / \textcolor{BrickRed}{0.696} & \textcolor{ForestGreen}{0.940} / \textcolor{BrickRed}{0.743} & 0.288 & \textcolor{ForestGreen}{0.418} / \textcolor{BrickRed}{0.397} & \textcolor{ForestGreen}{0.494} / \textcolor{BrickRed}{0.455} \\
                               & Llama3\_2-3b & 0.740 & \textcolor{ForestGreen}{0.747} / \textcolor{BrickRed}{0.541} & \textcolor{ForestGreen}{0.849} / \textcolor{BrickRed}{0.676}  & 0.714 & \textcolor{ForestGreen}{0.735} / \textcolor{BrickRed}{0.509} & \textcolor{ForestGreen}{0.855} / \textcolor{BrickRed}{0.652} \\
                               & Qwen2\_5-3b  & 0.772 & \textcolor{ForestGreen}{0.830} / \textcolor{BrickRed}{0.609} & \textcolor{ForestGreen}{0.856} / \textcolor{BrickRed}{0.641} &  0.797 & \textcolor{ForestGreen}{0.857} / \textcolor{BrickRed}{0.610} & \textcolor{ForestGreen}{0.874} / \textcolor{BrickRed}{0.642} \\
                               & Llama3\_1-8b & 0.817 & \textcolor{ForestGreen}{0.641} / \textcolor{BrickRed}{0.445} & \textcolor{ForestGreen}{0.863} / \textcolor{BrickRed}{0.648} & 0.832 & \textcolor{ForestGreen}{0.677} / \textcolor{BrickRed}{0.455} & \textcolor{ForestGreen}{0.847} / \textcolor{BrickRed}{0.596} \\
                               & Qwen2\_5-14b & 0.896 & \textcolor{ForestGreen}{0.968} / \textcolor{BrickRed}{0.790} & \textcolor{ForestGreen}{0.972} / \textcolor{BrickRed}{0.823}  & 0.897 & \textcolor{ForestGreen}{0.971} / \textcolor{BrickRed}{0.777} & \textcolor{ForestGreen}{0.973} / \textcolor{BrickRed}{0.800} \\
                               & Qwen2\_5-32b & 0.934 & \textcolor{ForestGreen}{0.972} / \textcolor{BrickRed}{0.759} & \textcolor{ForestGreen}{0.981} / \textcolor{BrickRed}{0.802} & 0.933 & \textcolor{ForestGreen}{0.968} / \textcolor{BrickRed}{0.750} & \textcolor{ForestGreen}{0.978} / \textcolor{BrickRed}{0.792} \\
                               & Yi-34b & 0.880 & \textcolor{ForestGreen}{0.888} / \textcolor{BrickRed}{0.629} & \textcolor{ForestGreen}{0.917} / \textcolor{BrickRed}{0.681} & 0.874 & \textcolor{ForestGreen}{0.868} / \textcolor{BrickRed}{0.582} & \textcolor{ForestGreen}{0.896} / \textcolor{BrickRed}{0.615} \\
                               & Llama3\_1-70b & 0.933 & \textcolor{ForestGreen}{0.938} / \textcolor{BrickRed}{0.649} & \textcolor{ForestGreen}{0.975} / \textcolor{BrickRed}{0.794} & 0.935 & \textcolor{ForestGreen}{0.945} / \textcolor{BrickRed}{0.690} & \textcolor{ForestGreen}{0.977} / \textcolor{BrickRed}{0.811} \\
                               & Qwen2\_5-72b & 0.941 & \textcolor{ForestGreen}{0.971} / \textcolor{BrickRed}{0.808} & \textcolor{ForestGreen}{0.981} / \textcolor{BrickRed}{0.875}  & 0.939 & \textcolor{ForestGreen}{0.962} / \textcolor{BrickRed}{0.750} & \textcolor{ForestGreen}{0.982} / \textcolor{BrickRed}{0.815} \\
                               \hline
                               
\multirow{9}{*}{USMLE} & phi3\_5 & 0.571 & \textcolor{ForestGreen}{0.781} / \textcolor{BrickRed}{0.612} & \textcolor{ForestGreen}{0.838} / \textcolor{BrickRed}{0.703} &  0.189 & \textcolor{ForestGreen}{0.319} / \textcolor{BrickRed}{0.306} & \textcolor{ForestGreen}{0.363} / \textcolor{BrickRed}{0.372} \\
                               & Llama3\_2-3b & 0.634 & \textcolor{ForestGreen}{0.618} / \textcolor{BrickRed}{0.428} & \textcolor{ForestGreen}{0.785} / \textcolor{BrickRed}{0.614} &  0.658 & \textcolor{ForestGreen}{0.596} / \textcolor{BrickRed}{0.407} & \textcolor{ForestGreen}{0.767} / \textcolor{BrickRed}{0.572} \\
                               & Qwen2\_5-3b & 0.477 & \textcolor{ForestGreen}{0.670} / \textcolor{BrickRed}{0.563} & \textcolor{ForestGreen}{0.740} / \textcolor{BrickRed}{0.648} &  0.514 & \textcolor{ForestGreen}{0.663} / \textcolor{BrickRed}{0.529} & \textcolor{ForestGreen}{0.740} / \textcolor{BrickRed}{0.630} \\
                               & Llama3\_1-8b & 0.627 & \textcolor{ForestGreen}{0.371} / \textcolor{BrickRed}{0.290} & \textcolor{ForestGreen}{0.637} / \textcolor{BrickRed}{0.507} &  0.679 & \textcolor{ForestGreen}{0.431} / \textcolor{BrickRed}{0.325} & \textcolor{ForestGreen}{0.730} / \textcolor{BrickRed}{0.585} \\
                               & Qwen2\_5-14b & 0.744 & \textcolor{ForestGreen}{0.897} / \textcolor{BrickRed}{0.699} & \textcolor{ForestGreen}{0.911} / \textcolor{BrickRed}{0.750} &  0.732 & \textcolor{ForestGreen}{0.898} / \textcolor{BrickRed}{0.690} & \textcolor{ForestGreen}{0.906} / \textcolor{BrickRed}{0.739} \\
                               & Qwen2\_5-32b & 0.811 & \textcolor{ForestGreen}{0.898} / \textcolor{BrickRed}{0.659} & \textcolor{ForestGreen}{0.923} / \textcolor{BrickRed}{0.744} &  0.816 & \textcolor{ForestGreen}{0.906} / \textcolor{BrickRed}{0.661} & \textcolor{ForestGreen}{0.931} / \textcolor{BrickRed}{0.744} \\
                               & Yi-34b & 0.652 & \textcolor{ForestGreen}{0.777} / \textcolor{BrickRed}{0.575} & \textcolor{ForestGreen}{0.831} / \textcolor{BrickRed}{0.666} &  0.652 & \textcolor{ForestGreen}{0.726} / \textcolor{BrickRed}{0.513} & \textcolor{ForestGreen}{0.781} / \textcolor{BrickRed}{0.572} \\
                               & Llama3\_1-70b & 0.885 & \textcolor{ForestGreen}{0.849} / \textcolor{BrickRed}{0.493} & \textcolor{ForestGreen}{0.946} / \textcolor{BrickRed}{0.679} &  0.886 & \textcolor{ForestGreen}{0.841} / \textcolor{BrickRed}{0.485} & \textcolor{ForestGreen}{0.941} / \textcolor{BrickRed}{0.666} \\
                               & Qwen2\_5-72b & 0.849 & \textcolor{ForestGreen}{0.930} / \textcolor{BrickRed}{0.668} & \textcolor{ForestGreen}{0.948} / \textcolor{BrickRed}{0.726} & 0.858 & \textcolor{ForestGreen}{0.909} / \textcolor{BrickRed}{0.645} & \textcolor{ForestGreen}{0.939} / \textcolor{BrickRed}{0.729} \\
                               \hline
                               
\multirow{9}{*}{CMCQRD} & phi3\_5 & 0.706 & \textcolor{ForestGreen}{0.818} / \textcolor{BrickRed}{0.664} & \textcolor{ForestGreen}{0.845} / \textcolor{BrickRed}{0.712} & 0.246 & \textcolor{ForestGreen}{0.374} / \textcolor{BrickRed}{0.369} & \textcolor{ForestGreen}{0.397} / \textcolor{BrickRed}{0.417} \\
                               & Llama3\_2-3b & 0.666 & \textcolor{ForestGreen}{0.671} / \textcolor{BrickRed}{0.493} & \textcolor{ForestGreen}{0.806} / \textcolor{BrickRed}{0.635} &  0.705 & \textcolor{ForestGreen}{0.688} / \textcolor{BrickRed}{0.518} & \textcolor{ForestGreen}{0.829} / \textcolor{BrickRed}{0.679} \\
                               & Qwen2\_5-3b & 0.662 & \textcolor{ForestGreen}{0.791} / \textcolor{BrickRed}{0.622} & \textcolor{ForestGreen}{0.816} / \textcolor{BrickRed}{0.659} & 0.744 & \textcolor{ForestGreen}{0.864} / \textcolor{BrickRed}{0.673} & \textcolor{ForestGreen}{0.884} / \textcolor{BrickRed}{0.704} \\
                               & Llama3\_1-8b & 0.752 & \textcolor{ForestGreen}{0.583} / \textcolor{BrickRed}{0.415} & \textcolor{ForestGreen}{0.800} / \textcolor{BrickRed}{0.602} & 0.781 & \textcolor{ForestGreen}{0.650} / \textcolor{BrickRed}{0.454} & \textcolor{ForestGreen}{0.828} / \textcolor{BrickRed}{0.628} \\
                               & Qwen2\_5-14b & 0.888 & \textcolor{ForestGreen}{0.95} / \textcolor{BrickRed}{0.737} & \textcolor{ForestGreen}{0.958} / \textcolor{BrickRed}{0.747} & 0.894 & \textcolor{ForestGreen}{0.951} / \textcolor{BrickRed}{0.74} & \textcolor{ForestGreen}{0.957} / \textcolor{BrickRed}{0.770} \\
                               & Qwen2\_5-32b & 0.908 & \textcolor{ForestGreen}{0.954} / \textcolor{BrickRed}{0.728} & \textcolor{ForestGreen}{0.970} / \textcolor{BrickRed}{0.785} &  0.912 & \textcolor{ForestGreen}{0.949} / \textcolor{BrickRed}{0.743} & \textcolor{ForestGreen}{0.969} / \textcolor{BrickRed}{0.813} \\
                               & Yi-34b & 0.841 & \textcolor{ForestGreen}{0.873} / \textcolor{BrickRed}{0.624} & \textcolor{ForestGreen}{0.900} / \textcolor{BrickRed}{0.668} & 0.831 & \textcolor{ForestGreen}{0.794} / \textcolor{BrickRed}{0.546} & \textcolor{ForestGreen}{0.831} / \textcolor{BrickRed}{0.582} \\
                               & Llama3\_1-70b & 0.917 & \textcolor{ForestGreen}{0.926} / \textcolor{BrickRed}{0.684} & \textcolor{ForestGreen}{0.973} / \textcolor{BrickRed}{0.809} & 0.912 & \textcolor{ForestGreen}{0.912} / \textcolor{BrickRed}{0.617} & \textcolor{ForestGreen}{0.972} / \textcolor{BrickRed}{0.789} \\
                               & Qwen2\_5-72b & 0.914 & \textcolor{ForestGreen}{0.962} / \textcolor{BrickRed}{0.737} & \textcolor{ForestGreen}{0.971} / \textcolor{BrickRed}{0.766} & 0.912 & \textcolor{ForestGreen}{0.955} / \textcolor{BrickRed}{0.739} & \textcolor{ForestGreen}{0.973} / \textcolor{BrickRed}{0.793} \\
\hline
\end{tabular}
\end{table*}

\section{Large Language Models Used}
\label{app:LLMs_used}
Table \ref{tab:LLMs_used} presents the collection of Large Language Models used in our experiments involving model uncertainty. 

\begin{table*}[h!]
    \centering
    \renewcommand{\arraystretch}{1.2}  
    \caption{\textbf{Default precision and quantised LLMs used in the experiments. All models can be found at the HuggingFace Hub.}}
    \label{tab:LLMs_used}
    \begin{tabular}{lll}\hline
        Model Name          & Default Precision   & 4-bit Quantisation                                                           \\
        \hline
        phi3\_5-chat        & \url{microsoft/Phi-3.5-mini-instruct}    &  \url{unsloth/Phi-3.5-mini-instruct-bnb-4bit}        \\
        Llama3\_2-3b-chat   & \url{meta-llama/Llama-3.2-3B-Instruct}   &  \url{unsloth/Llama-3.2-3B-Instruct-bnb-4bit}        \\
        Qwen2\_5-3b-chat    & \url{Qwen/Qwen2.5-3B-Instruct}           &  \url{unsloth/Qwen2.5-3B-Instruct-bnb-4bit}          \\
        Llama3\_1-8b-chat   & \url{meta-llama/Llama-3.1-8B-Instruct}   &  \url{unsloth/Meta-Llama-3.1-8B-Instruct-bnb-4bit}   \\
        Qwen2\_5-14b-chat   & \url{Qwen/Qwen2.5-14B-Instruct}          &  \url{unsloth/Qwen2.5-14B-Instruct-bnb-4bit}         \\        
        Qwen2\_5-32b-chat   & \url{Qwen/Qwen2.5-32B-Instruct}          &  \url{unsloth/Qwen2.5-32B-Instruct-bnb-4bit}         \\
        Yi-34b-chat         & \url{01-ai/Yi-34B-Chat}                  &  \url{unsloth/yi-34b-chat-bnb-4bit}                  \\
        Llama3\_1-70b-chat  & \url{meta-llama/Llama-3.1-70B-Instruct}  &  \url{unsloth/Meta-Llama-3.1-70B-Instruct-bnb-4bit}  \\
        Qwen2\_5-72b-chat   & \url{Qwen/Qwen2.5-72B-Instruct}          &  \url{unsloth/Qwen2.5-72B-Instruct-bnb-4bit}         \\
        \hline 
    \end{tabular}%
\end{table*}

\section{Results Overview}
\label{app:all_results}

Table \ref{tab:app:all_results} presents an overview of the results of all experiments, along with Table \ref{tab:app:Bio_Clinical_BERT} which presents the performance our setup using the Bio\_clinical Bert Embeddings \citep{alsentzer-etal-2019-publicly}. The results using BERT embeddings are repeated to facilitate comparison. In our experiments, using the Bio\_clinical Bert Embeddings consistently led to worse performance (higher RMSE) for all experiments.

\begin{table*}
\centering
\renewcommand{\arraystretch}{1.2}  
\caption{\textbf{Root Mean Squared Error (RMSE, the lower the better) on the test set using different sets of features. Best overall results are highlighted in \hlgreen{light green}. All results are averaged over ten repetitions, with the standard deviation not exceeding 0.002.}}
\label{tab:app:all_results}
\begin{tabular}{lcc:cc:cc:cc} 
\hline
\textbf{Method} & \multicolumn{2}{c}{\textbf{Biopsychology}} & \multicolumn{2}{c}{\textbf{USMLE}} & \multicolumn{2}{c}{\textbf{CMCQRD}} & \multicolumn{2}{c}{\textbf{CMCQRD\_IRT}} \\  
\hline
\multicolumn{7}{l}{\hlgray{\textbf{Baselines}}} \\
Dummy Regressor                                             & \multicolumn{2}{c}{0.1667}  & \multicolumn{2}{c}{0.3110}  & \multicolumn{2}{c}{0.1833} & \multicolumn{2}{c}{9.7568}      \\
Best Literature Result \citep{raina-etal-2025-finetuning}                                      & \multicolumn{2}{c}{-}       & \multicolumn{2}{c}{0.291}  & \multicolumn{2}{c}{-} & \multicolumn{2}{c}{8.5}\\
Linguistic Features Baseline \citep{ha-etal-2019-predicting} & \multicolumn{2}{c}{0.1544} & \multicolumn{2}{c}{0.3147}  & \multicolumn{2}{c}{0.1852} & \multicolumn{2}{c}{9.3335}      \\
\hline
\multicolumn{7}{l}{\hlgray{\textbf{Only Text}}} \\
TF-IDF                                                      & \multicolumn{2}{c}{0.1479}  & \multicolumn{2}{c}{0.3092}  & \multicolumn{2}{c}{0.1843} & \multicolumn{2}{c}{9.2531}      \\
BERT Embeddings                                             & \multicolumn{2}{c}{0.1498}  & \multicolumn{2}{c}{0.3066}  & \multicolumn{2}{c}{0.1843} & \multicolumn{2}{c}{8.9347}      \\  
\hline
\multicolumn{7}{l}{\hlgray{\textbf{Only Uncertainty}}} \\
                                                            & \textbf{Default}   & \textbf{4-bit}  & \textbf{Default} & \textbf{4-bit}   & \textbf{Default}      & \textbf{4-bit}      & \textbf{Default} & \textbf{4-bit}            \\
1st Token Probabilities                                     & 0.1473          & 0.1539	        & 0.3041        & 0.2960	       & 0.1770             & 0.1752              & 9.4161        & 9.5385                    \\
Choice Order Sensitivity                                    & 0.1543          & 0.1582	        & 0.3155        & 0.3178	       & 0.1788             & 0.1880              & 9.9787        & 9.8563                    \\
Both Uncertainty Features                                   & 0.1460          & 0.1538	        & 0.3034        & 0.2968	       & 0.1754             & 0.1761              & 9.3705        & 9.4899                    \\  
\hline
\multicolumn{7}{l}{\hlgray{\textbf{Only Choice Similarity}}} \\
General (\texttt{all-MiniLM-L6-v2})                        & \multicolumn{2}{c}{0.1895}         & \multicolumn{2}{c}{0.3567}       & \multicolumn{2}{c}{0.2226}               & \multicolumn{2}{c}{12.2829}               \\
Medical (\texttt{S-PubMedBert-MS-MARCO})                   & \multicolumn{2}{c}{0.1883}         & \multicolumn{2}{c}{0.3432}       & \multicolumn{2}{c}{0.2146}               & \multicolumn{2}{c}{11.4768}               \\  
\hline
                                                            & \textbf{TF-IDF}   & \textbf{BERT}  & \textbf{TF-IDF} & \textbf{BERT}   & \textbf{TF-IDF}      & \textbf{BERT}      & \textbf{TF-IDF} & \textbf{BERT}        \\
\multicolumn{7}{l}{\hlgray{\textbf{Text and Uncertainty}}} \\
First Token Probability (Default)                              & 0.1361            & 0.1362         & 0.3037          & 0.2868          & 0.1668               & 0.1669             & 8.5661          & 8.2763               \\
Choice Order Sensitivity (Default)                             & 0.1378            & 0.1409         & 0.3100          & 0.2906          & 0.1658               & 0.1676             & 8.6663          & 8.4204               \\
Both Uncertainty Features (Default)                            & 0.1359            & 0.1365         & 0.3044          & 0.2864          & 0.1654               & 0.1669             & 8.5933          & 8.3009               \\  
First Token Probability (4-bit)                             & 0.1365            & 0.1385         & 0.2851          & 0.2854          & 0.1680               & 0.1675             & 8.5392          & 8.3322               \\
Choice Order Sensitivity (4-bit)                           & \hlgreen{0.1309}    & 0.1411         & 0.2951          & 0.2961          & 0.1671               & 0.1672             & 8.6218          & 8.3459               \\
Both Uncertainty Features (4-bit)                           & 0.1371            & 0.1388         & 0.2856          & 0.2846          & 0.1685               & 0.1670             & 8.5587          & 8.3728               \\  
\hline
\multicolumn{7}{l}{\hlgray{\textbf{Text, Both Uncertainties \& Choice Similarity}}} \\
General Sim. (Default)                                         & 0.1372            &  0.1367        &  0.2835         & 0.2862          & \hlgreen{0.1651}   & 0.1669             & 8.5934          &  8.2956                 \\
Medical Sim. (Default)                                         & 0.1376            &  0.1361        &  0.2836         & 0.2853          & \hlgreen{0.1651}   & 0.1664             & 8.5999          &  \hlgreen{8.2325}        \\
Both Sim. (Default)                                            & 0.1378            &  0.1365        &  0.2847         & 0.2862          & 0.1653            & 0.1676             & 8.5694          &  8.2817                 \\  
General Sim. (4-bit)                                        & 0.1386            &  0.1389        &  0.2850         & 0.2841          & 0.1674            & 0.1659             & 8.5696          &  8.3338                 \\
Medical Sim. (4-bit)                                        & 0.1378            &  0.1381        &  0.2856         & 0.2844          & 0.1676            & 0.1660             & 8.5204          &  8.2801                 \\
Both Sim. (4-bit)                                           & 0.1397            &  0.1412        &  0.2860         & 0.2850          & 0.1673            & 0.1659             & 8.5475          &  8.2820                 \\  
\hline
\multicolumn{7}{l}{\hlgray{\textbf{Text, Both Uncertainties, Both Sim \& Linguistic Features}}}  \\
 Default      & 0.1393            &  0.1362        &  \hlgreen{0.2817}            & 0.2857          & 0.1652               & 0.1673             & 8.5490          &  8.6575             \\
 4-bit     & 0.1411            &  0.1410        &  0.2853                     & 0.2844          & 0.1677               & 0.1660             & 8.4966          &  8.5907             \\  
\hline
\end{tabular}
\end{table*}

\begin{table*}
\centering
\renewcommand{\arraystretch}{1.2}  
\caption{\textbf{Comparison in Root Mean Squared Error (RMSE, the lower the better) achieved using BERT or Bio\_Clinical BERT embeddings. Experiments using the CMCQRD were not conducted, as the dataset does not cover biological/medical domain. All results are averaged over ten repetitions, with the standard deviation not exceeding 0.002.}}
\label{tab:app:Bio_Clinical_BERT}
\begin{tabular}{lcc:cc} 
\hline
\multirow{2}{*}{\textbf{Method}} & \multicolumn{2}{c}{\textbf{Biopsychology}} & \multicolumn{2}{c}{\textbf{USMLE}}\\  
\cmidrule(lr){2-3} \cmidrule(lr){4-5}
                                                            & \textbf{BERT}   & \textbf{Bio\_Clinical BERT}  & \textbf{BERT} & \textbf{Bio\_Clinical BERT}     \\
\hline
\multicolumn{5}{l}{\hlgray{\textbf{Text and Uncertainty}}} \\
First Token Probability (Default)                              & 0.1362          & 0.1376                       & 0.2868            &  0.2945                     \\
Choice Order Sensitivity (Default)                             & 0.1409          & 0.1430                       & 0.2906            &  0.3004                     \\
Both Uncertainty Features (Default)                            & 0.1365          & 0.1387                       & 0.2864            &  0.2952                     \\  
First Token Probability (4-bit)                             & 0.1385          & 0.1396                       & 0.2854            &  0.2908                     \\
Choice Order Sensitivity (4-bit)                            & 0.1411          & 0.1426                       & 0.2961            &  0.3035                     \\
Both Uncertainty Features (4-bit)                           & 0.1388          & 0.1392                       & 0.2846            &  0.2909                     \\  
\hline   
\multicolumn{5}{l}{\hlgray{\textbf{Text, Both Uncertainties \& Choice Similarity}}} \\   
General Sim. (Default)                                         & 0.1367          & 0.1380                       & 0.2862        &  0.2951                           \\
Medical Sim. (Default)                                         & 0.1361          & 0.1381                       & 0.2853        &  0.2936                           \\
Both Sim. (Default)                                            & 0.1365          & 0.1377                       & 0.2862        &  0.2936                           \\  
General Sim. (4-bit)                                        & 0.1389          & 0.1424                       & 0.2841        &  0.2926                           \\
Medical Sim. (4-bit)                                        & 0.1381          & 0.1425                       & 0.2844        &  0.2924                           \\
Both Sim. (4-bit)                                           & 0.1412          & 0.1422                       & 0.2850        &  0.2922                           \\  
\hline   
\multicolumn{5}{l}{\hlgray{\textbf{Text, Both Uncertainties, Both Sim \& Linguistic Features}}}  \\   
 Default                                                       & 0.1362          & 0.138                        &  0.2857       &  0.2939                         \\
 4-bit                                                      & 0.1410          & 0.1426                       &  0.2844       &  0.2915                         \\  
\hline
\end{tabular}
\end{table*}

\end{document}